\documentclass[final]{cvpr}

\usepackage{times}
\usepackage{epsfig}
\usepackage{graphicx}
\usepackage{amsmath}
\usepackage{amssymb}
\usepackage{algorithm}
\usepackage{algpseudocode}
\usepackage{soul}
\usepackage{caption}
\usepackage{subcaption}

\usepackage[pagebackref=true,breaklinks=true,letterpaper=true,colorlinks,bookmarks=false]{hyperref}

\def\cvprPaperID{8121} 

\DeclareMathOperator*{\argmax}{arg\,max}
\newcommand{\cS}{\mathcal{S}}
\newcommand{\cB}{\mathcal{B}}
\newcommand{\cA}{\mathcal{A}}
\newcommand{\cI}{\mathcal{I}}
\newcommand{\cX}{\mathcal{X}}
\newcommand{\cM}{\mathcal{M}}

\newcommand{\cV}{\mathcal{V}}

\newcommand{\bv}{\mathbf{v}}
\newcommand{\bS}{\mathbf{S}}
\newcommand{\bH}{\mathbf{H}}

\newcommand{\bb}{\mathbf{b}}
\newcommand{\ba}{\mathbf{a}}

\newcommand{\bx}{\mathbf{x}}
\newcommand{\btheta}{\boldsymbol{\theta}}

\newcommand{\bbR}{\mathbb{R}}
\begin{document}

\title{Unsupervised Learning for Robust Fitting:\\ A Reinforcement Learning Approach}

\author{{Giang Truong* \quad Huu Le** \quad David Suter* \quad Erchuan Zhang* \quad Syed Zulqarnain Gilani*
}\\
*School of Science, Edith Cowan University, Australia\\
**Department of Electrical Engineering, Chalmers University of Technology\\
{\tt\small {\{h.truong, d.suter, erchuan.zhang, s.gilani\}}@ecu.edu.au, 
\tt\small {huul}@chalmers.se}
}

\maketitle

\begin{abstract}
Robust model fitting is a core algorithm in a large number of computer vision applications. Solving this problem efficiently for datasets highly contaminated with outliers is, however, still challenging due to the underlying computational complexity. Recent literature has focused on learning-based algorithms.
However, most approaches are supervised which require a large amount of labelled training data. In this paper, we introduce a novel unsupervised learning framework that learns to directly solve robust model fitting. Unlike other methods, our work is agnostic to the underlying input features, and can be easily generalized to a wide variety of LP-type problems with quasi-convex residuals. We empirically show that our method outperforms existing unsupervised learning approaches, and achieves competitive results compared to traditional methods on several important computer vision problems\footnote{Our source code will be made available}. 
\end{abstract}

\vspace{-3mm}
\section{Introduction}
\vspace{-2mm}
Many computer vision applications require the estimation of a model from a set of observations~\cite{hartley2003multiple}.
In outlier-free settings, fitting a geometric model to a dataset can be performed relatively easily by, for example, solving a least squares problem. However, in the presence of outliers in the data, a robust estimator~\cite{fischler1981random,huber81} must be employed to ensure the stable performance of any algorithm. 
As an example, consider SLAM~\cite{mur2015orb}, which is now a fundamental building block in several robotic or autonomous driving pipelines. It requires multiple estimations of the fundamental/ essential matrices between the consecutive views captured along the camera trajectory. In many circumstances, erroneous correspondences between the frames could lead to incorrect camera pose estimation. Consequently, if the outliers are not removed, the whole tracking trajectory could be severely affected. Therefore, it is desirable to design robust fitting algorithms that are highly accurate and able to achieve real-time performance. This is a challenging task, as solving robust fitting optimally has been shown to be NP-hard~\cite{chin2018robust,chin2017maximum}.

\begin{figure}[t]
    \centering
    \includegraphics[width=1\linewidth]{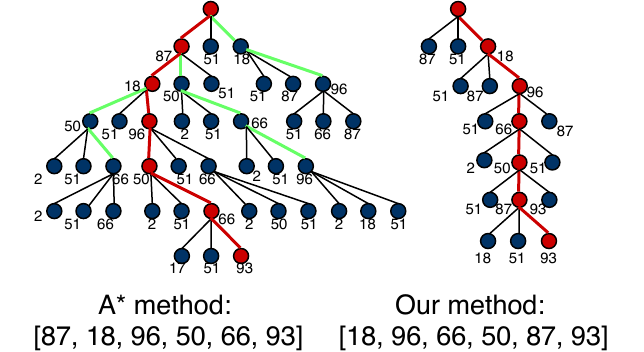}
    \caption{Illustration of the solutions found by our unsupervised learning method (right) and a globally optimal algorithm~\cite{chin15} (left). The number shows the specific index of points in the point set. The admissible heuristic in A* method brings the search into some fruitless subparts (green line) before discovering optimal solution (red line). Our agent learns to remove outliers by traversing from the initial state to the goal state in the minimal number of steps (the states are numbered based on the index of the removed point). Observe that both methods terminate at the same solution (i.e., both remove the same set of outliers).}
    \label{fig:coverPic}
    \vspace{-5mm}
\end{figure}

In addition to popular methods such as Random Sample Consensus (RANSAC)~\cite{fischler1981random} and a number of randomized or deterministic  variants~\cite{chum2003locally,chum2005matching,loransac, le2017exact, cai2018deterministic, chin15, cai2019consensus}, the advent of deep learning in recent years has inspired research in learning-based approaches for robust estimation~\cite{ranftl2018deep,sarlin2020superglue,moo2018learning, probst2019unsupervised,detone2016deep,le2020deep}. The main idea behind these techniques is to exploit the learning capabilities of deep Convolutional Neural Networks (CNNs) to directly regress the robust estimates~\cite{le2020deep,detone2016deep}, or quickly identify the outliers~\cite{moo2018learning} 
These approaches have demonstrated their superior performance on many datasets, and hence, developing learning-based robust estimators can be a promising research direction. 
However, most learning techniques mentioned above are supervised, hence they typically require a large amount of labelled data. This potential bottleneck could be resolved by either generating ground truth data automatically by using synthetic data, or by using conventional methods (e.g. RANSAC) to generate ground truth. However, these "quick fixes" have their own drawbacks. 
Specifically, a network fully trained on synthetic data may not be able to generalize well to real world scenarios since it has not been exposed to real examples during training. 
Similarly, ground truth obtained from classic conventional methods is not guaranteed to be the gold standard as the obtained solutions may be incorrect. One could, on the other hand, employ some global consensus maximization methods~\cite{chin15} to generate the ground truth, but this would be at the cost of an exceptionally slow training process. Moreover, some methods are problem-specific, and it is non-trivial to extend them to other robust fitting tasks.


We address these problems and present a novel unsupervised learning framework for robust fitting.
Inspired by the success of Reinforcement Learning (RL) in several unsupervised tasks~\cite{mnih2013playing,sutton2018reinforcement}, we cast our robust fitting problem as a special case of goal oriented learning. Such a transformation is achieved thanks to the underlying tree structure of consensus maximization~\cite{chin15,le2017ratsac,cai2019consensus}.
Moreover, we also propose a novel architecture that efficiently captures the instantaneous state of the data during transition. 
Fig.~\ref{fig:coverPic} shows an example of a 2D line fitting problem, where we plot the trajectory of A*~\cite{chin15} (a globally optimal algorithm), and the path traversed by our agent from the initial state to the goal state. Observe that both remove the same set of outliers, which demonstrates the learning capability of our network to effectively explore the environment. Furthermore, in contrast to the implementation of A*, which explores redundant bases before reaching the optimal, our network can quickly identify the shortest path to reach the goal state, resulting in significantly faster run times. (see Section.~\ref{sec:method} for more detail.)
To the best of our knowledge, our work is the first to learn a deep architecture model in a reinforcement learning paradigm for consensus maximization in computer vision.

\vspace{-4mm}
\paragraph{Contributions} The main contributions of our paper can be summarized as follows:
\vspace{-2mm}
\begin{itemize}
    \item We propose a novel unsupervised learning framework for robust estimation. By exploiting the \textit{special structure} (see Section.~\ref{sec:method}) of the problem under the consensus maximization formulation, we incorporate robust model fitting into the well-known goal-oriented reinforcement learning framework, resulting in an efficient learning mechanism without any supervision.
    \vspace{-2mm}
    \item We develop a new state embedding scheme  based on a graph convolutional network, and an efficient back-bone network that allows our agent to effectively explore the action space to achieve the goal state. (see Section.~\ref{sec:state_encoding} and Section.~\ref{sec:agent_design})
\end{itemize}
\vspace{-3mm}
\subsection{Related Work}
\vspace{-2mm}
Solving robust model fitting has a rich literature, where the randomized method such as RANSAC~\cite{fischler1981random} is considered to be the most popular approach because it is simple to  implement, and provides competitive results for many real world problems. The sampling scheme behind RANSAC has inspired many of its variants~\cite{loransac, torr2000mlesac,chum2005matching}, but using them on  data highly contaminated with outliers  still results in unsatisfactory outcomes. In addition to the randomized variants whose purpose is to improve RANSAC, deterministic schemes have also become popular~\cite{olsson2010outlier, olsson07, le2017alternating, cai2018deterministic}. However, their use in real-time applications is still limited due to their long processing time. On the other hand, algorithms that offer globally optimal solutions are also under active research~\cite{chin15, cai2019consensus}. Despite their elegant algorithmic constructions, which often do improve their run time, their general long run time renders them impractical to use them for most real-world applications. 

Learning to fit robust models is an interesting idea that has emerged in recent years, thanks to the learning capability of deep CNNs. Such methods have shown promising results in several fitting tasks. Ranftl et al.~\cite{ranftl2018deep} proposed a deep network for fundamental matrix estimation by learning the weights for the residuals. Learning to solve homography estimation has also been addressed in~\cite{le2020deep}. The drawback of these methods is that they are problem-specific, hence extending to a more general class of model fitting instances is not very trivial. The use of an attention mechanism for robust feature matching has also been considered~\cite{moo2018learning, sarlin2020superglue}. The idea behind these approaches is to learn to classify correct inliers from the potentially contaminated set. Although efficient, most learning-based approaches require a sufficiently large amount of training data, which could make them impractical in new environments.

Our work is closely related to the unsupervised learning approach for consensus maximization proposed by Probst et al.~\cite{probst2019unsupervised}. However, we take a different approach by exploiting the tree structure of the underlying fitting problem. This allows us to easily train our network from scratch, while~\cite{probst2019unsupervised} requires a certain supervised signal in order to work effectively. We show that training our network from scratch in an unsupervised fashion is straightforward and at the same time the results are competitive  when compared with~\cite{probst2019unsupervised}. 



\vspace{-3mm}
\section{Background}
\vspace{-1mm}
\subsection{Problem Formulation}
\vspace{-2mm}
While there exist different ways to formulate robust estimates, our work uses the popular consensus set maximization formulation~\cite{chin15}.  
The objective is to find an estimate $\btheta^* \in \bbR^d$ that is consistent with as many of the observations $\cX=\{\bx_{i}\}_{i=1}^N$ as possible, i.e., finding the maximum number of inliers up to a predefined threshold $\epsilon$:
\vspace{-3mm}
\begin{equation}
\begin{aligned}
& \max_{\btheta \in \bbR^d, \cI \subseteq \cX} 
& & | \cI |  \\
& \text{subject to}
& & r(\bx_{i}|\btheta) \leq \epsilon, \forall \bx_{i} \in \cX.
\end{aligned}
\label{eq:maxcon}
\end{equation}
where $r(\bx_{i}|\btheta)$ is the residual of $\bx_{i}$ w.r.t model parameter $\btheta$. 
The solution $(\cI^*, \btheta^*)$ of~\eqref{eq:maxcon} provides the optimal inlier set $\cI^*$ that is consistent with the estimate $\btheta^*$. 
Similar to other consensus maximization approaches, we also focus on quasi-convex residuals having the form~\cite{ke05},
\vspace{-2mm}
\begin{equation}
r(\btheta) = \frac{p(\btheta)}{q(\btheta)},
\vspace{-2mm}
\end{equation}
where $p(\btheta) \ge 0$ is a convex function and $q(\btheta) > 0$ is a linear function. Objectives of several vision problems possess such quasi-convexity~\cite{ke05}. 
In this paper, to assist visualization of the mathematical concepts, most examples and analyses will be based on the linear fitting problems whose residual functions are in fact convex and can be written as follows: given $\bx_i = (\ba_i; b_i)$ with $\ba_i \in \bbR^d$ and $b_i \in \bbR$,
\begin{equation}
r(\btheta) = |\ba^T_{i}\btheta - b_{i}|.
\vspace{-2mm}
\end{equation}




\vspace{-2mm}
\subsection{Goal-Oriented Reinforcement Learning}
\vspace{-2mm}
Our proposed unsupervised learning approach for robust estimation is based upon the well-known goal-oriented reinforcement learning (RL) framework~\cite{sutton2018reinforcement}, which is briefly outlined in this subsection. This framework consists of an agent $A$ that aims to navigate in an environment to reach a pre-defined goal in the smallest number of steps to maximize a total reward. Each time step is associated with a state $s^{(t)}$. The agent can select an action $a^{(t)}_i \in \cA^{(t)}$, where $\cA^{(t)}$ is the set of actions that are available at time $t$. 
Based on the action $a$ taken by the agent, the environment will transition to a new state $s^{(t + 1)}$ and return a reward $r_{t}(a)$, where the reward function $r_t(.)$ depends on the particular application. The goal of the framework is for the agent to reach the pre-defined goal that maximizes the cumulative reward (also known as return),
\vspace{-5mm}
\begin{equation}
R = \sum_{t=t_0}^{\infty} \beta^{t - t_0} r_t,
\vspace{-2mm}
\end{equation}

obtained from the initial state to the final state, where $\beta^{t-t_0}$ is the discount factor to weigh the importance of the $Q$ value at a particular time step. 
Commonly, deep Q learning~\cite{mnih2013playing} is used in most RL frameworks, for the agent to learn the optimal actions. In particular, Q learning is a model-free RL method which learns the quality of actions for the agent to take appropriate action under a particular circumstance. Deep Q-learning is the fundamental model used in our work, and will be further outlined in Section.~\ref{sec:method}

\vspace{-3mm}
\section{Proposed Unsupervised Learning Approach}
\label{sec:method}
\vspace{-2mm}
While RL has shown its strength in several applications, applying it to robust estimation is by no means a trivial task. The main challenges lie in the definition of a state, reward function, goal specification and designing of an agent that can learn to efficiently explore the environment in an optimal way.  In this section, we introduce a novel framework that enables the use of RL for our robust fitting problem. 

\vspace{-2mm}
\subsection{Definitions}
\label{Sec:Definitions}

\vspace{-2mm}
\begin{figure}
    \centering
    \includegraphics[width = 0.6\columnwidth]{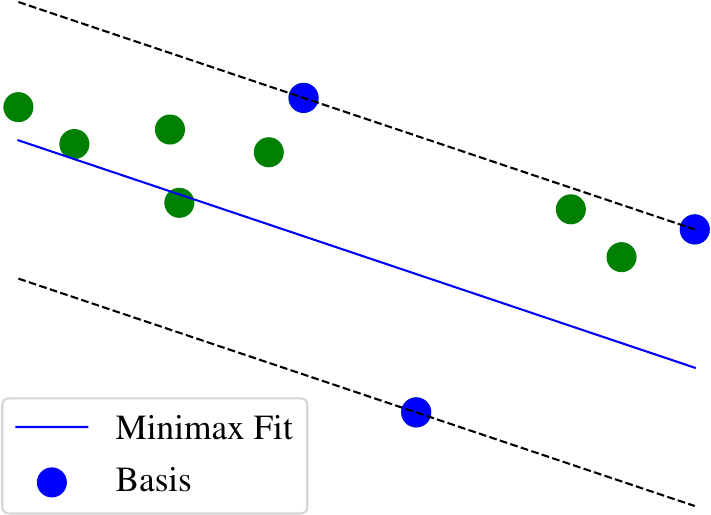}
    \caption{Illustration of a minimax fitting problem for a set of points in 2D. Blue dots represent measurements with largest residual, which form the \emph{basis set}.}
    \label{fig:minimax_2d}
    \vspace{-5mm}
\end{figure}

Given a set of measurements (observations) $\cX = \{\bx_i\}_{i=1}^N$, let us first consider the minimax fitting problem that returns the estimate which minimizes the maximum residual:
\vspace{-2mm}
\begin{equation}
    f(\cX)=\min_{\btheta \in \bbR^d, \gamma \in R} \gamma, \;\; \text{s.t.} \; r(\bx_i|\btheta) \le \gamma \;\; \forall i, \bx_i \in \cX.
    \label{eq:minimax}
    \vspace{-2mm}
\end{equation}
It can be proven that if the residual function $r(\bx_i|\btheta)$ is quasi-convex, the above problem is also quasi-convex, hence it can be solved efficiently up to global optimality using any off-the-shelf solver~\cite{ke05}. Fig.~\ref{fig:minimax_2d} shows an example of the minimax fit of a set of points in 2D. This problem is the core sub-problem in our learning framework.

Observe that if the optimal solution $\gamma^*$ obtained from solving the above problem is not greater than the inlier threshold, i.e., $\gamma^* \le \epsilon$, then the solution $\btheta^*$ obtained from~\eqref{eq:minimax} also solves the robust fitting problem~\eqref{eq:maxcon} (there are no outliers). Otherwise, the optimal consensus must be a subset $\cI^*$ such that $f(\cI^*) \le \epsilon$. Therefore, the goal of our agent is to gradually remove a subset of outliers to reach the subset $\cI^*$. Obviously, in order to maximize $\cI^*$, the number of outliers removed by our agent must be minimized.
\vspace{-5mm}
\paragraph{State.} Under the RL framework, let us consider associating each subset $\cS \subseteq \cX$ with a state $s_{\cS}$ (the detailed construction of states for our network will be discussed in the following sections), and set $s_{\cX}$ to the initial state at which our agent will start the exploration process, i.e., $s^{(t=0)} = s_{\cX}$. From the above discussion, at a particular state $s_{\cS}$, if the action taken by our agent is to remove one data point $\bx_j \in \cS$ such that $f(\cS \setminus {\bx_j}) < f(\cS)$, then the task of the agent is to find the state $s_{\cI}$ such that $f(\cI) \le \epsilon$ in the smallest number of steps (i.e., to minimize the number of outliers removed). Refer to Fig.~\ref{fig:stateDesign} for a visualization of a state and its associated actions for a 2D fitting problem.
\vspace{-5mm}
\paragraph{Goal and Reward Function.} We therefore define our goal state, based on the RL framework discussed in the previous section, as a state $s_{\cS}$ such that $f(\cS) \le \epsilon$. Also, we can now define a reward function $e(.)$ associated with a state $s_{\cS}$ to be
\vspace{-2mm}
\begin{equation}
\vspace{-2mm}
    e(s_{\cS}) = \begin{cases}
    \begin{aligned}
    & ~0 && \text{if} \;\; f(\cS) \le \epsilon\\
    & -1 && \text{otherwise.} \\
    \end{aligned}
    \end{cases}
    \label{eq:reward}
    \vspace{-2mm}
\end{equation}
As can be observed, the maximized total reward obtained by our agent if using reward function~\eqref{eq:reward} corresponds to moving from the initial state to the goal state in the minimum number of steps.
\vspace{-2mm}
\subsection{Generating Action Sets}
\label{sec:generating_actions}
\vspace{-2mm}
As previously discussed, the available actions associated with a particular state $s_{\cS}$ corresponds to the removal of points $\bx_j \in \cS$ such that $f(\cS \setminus {\bx_j}) < f(\cS)$. One could test all the points in $\cS$ to generate the action set. However, such exhaustive testing turns out to be unnecessary for our problem.
To reduce the number of available actions at each state, we exploit a special property of our application as follows. 

Assume $\btheta_{\cS}$ is the solution of~\eqref{eq:minimax} for a set $\cS$. Let us consider the set $\cB_{\cS}$ containing points having the largest residuals,
\vspace{-2mm}
\begin{equation}
    \cB_{\cS} = \{\bx_j \in \cS | r(\bx_j | \btheta_{\cS})= f(\cS) \}
    \vspace{-2mm}
\end{equation}
Following the terminology in \cite{chin15, cai2019consensus}, 
we also call $\cB_{\cS}$ a basis of the set $\cS$. In the example shown in Figure~\ref{fig:minimax_2d}, the basis for a 2D-line fitting problems w.r.t. to the current estimate are the points plotted in blue.

Given $\cB_{\cS}$, one can prove (see \cite{chin15, cai2019consensus}) that, 
\vspace{-2mm}
\begin{equation}
    f(\cS \setminus \bx_j)  < f(\cS), \;\; \forall \bx_j \in \cB_{\cS}.
    \vspace{-2mm}
\end{equation}

\begin{figure}[t]
    \centering
    \includegraphics[width=0.8\linewidth]{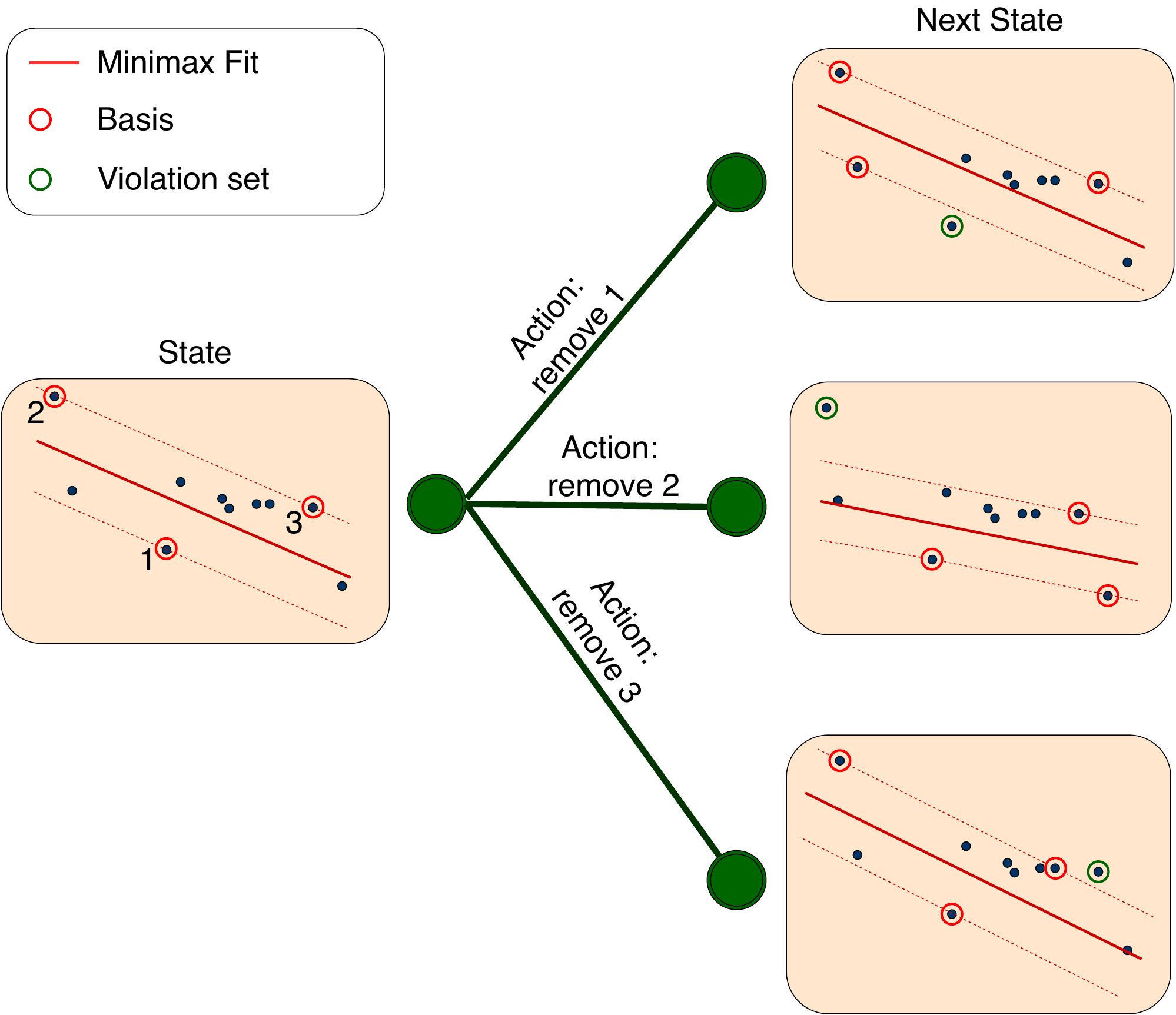}
    \caption{Visualization of states and actions. The initial state corresponds to the minimax fit of the original point set. The action, associated a particular state, consists of removing a point in the basis set and conduct minimax fit for the remaining points.}
    \label{fig:stateDesign}
    \vspace{-3mm}
\end{figure}

Intuitively, removing a point belonging to the basis set guarantees the reduction of the minimax fit for the remaining set (see Figure~\ref{fig:stateDesign}). This suggests that the actions associated with a state $s_{\cS}$ corresponds to the removal of a point in $\cB_{\cS}$ and conduct minimax fit for the remaining points. Figure~\ref{fig:stateDesign} visually illustrates how the actions can be generated from a particular state.

Moreover, for quasi-convex residuals in $d$ dimension, the maximum size of $\cB_{\cS}$ is proven to be $|\cB_{\cS}| \le d+1$( see ~\cite{eppstein2005quasiconvex}). 
Therefore, at a particular state, the maximum number of available actions in our cases is $d+1$. This property significantly reduces the action space for our learning framework, and is one of the key factors that leads to the efficacy of our learning scheme.  

\vspace{-2mm}
\subsection{State Encoding} 
\label{sec:state_encoding}
\vspace{-1mm}
In order to use the states above as the inputs to our network, they first need to be properly encoded.
Based on the above discussion, the crucial information for a state $s_{\cS}$ consist of the point set $\cS$ together with its basis $\cB_{\cS}$ (which also stores the information about the available actions associated with $s_{\cS}$). To enrich the information for $s_{\cS}$, our state encoding also comprises the set $\cV_{\cS} = \cX \setminus \cS$. Clearly, $\cV_{\cS}$ encodes the agent's state traversal history before reaching the state $s_{\cS}$. Given $\cS$, $\cB_{\cS}$ and $\cV_{\cS}$, we construct a matrix $\bS_{\cS}$ to feed it into our network. 

To encode $\cB_{\cS}$ and $\cV_{\cS}$, we define two binary vectors $\bb_{\cS} \in \{-1, 1\}^N$ and $\bv_{\cS} \in \{-1, 1\}^N$, respectively, defined as follows (we use the notation $\bx[i]$ to denote the $i$th component of a vector $\bx$),
\begin{equation}
    \bb_{\cS} [i] = 
    \begin{cases}
    \begin{aligned}
        1 &   &&\text{if} \; \bx_i \in \cB_{\cS} \\
        -1 & &&\text{otherwise}.
    \end{aligned}
    \end{cases} \text{and} \;
    \bv_{\cS} [i] = 
    \begin{cases}
    \begin{aligned}
        1  & \;  &&\text{if} \; \bx_i \in \cV_{\cS} \\
        -1 & \; &&\text{otherwise}.
    \end{aligned}
    \end{cases}
\end{equation}
Therefore, each state $s_{\cS}$ can now be encoded by the matrix  $\bS_{\cS}$,
\vspace{-2mm}
\begin{equation}
    \bS_{\cS} = \left[\bH \;\; \bb_{\cS} \;\; \bv_{\cS} \right],
    \vspace{-2mm}
\end{equation}
where $H$ is the matrix that collects all the data points in the input set $\cX$. More specifically, we set the $i$-th row of the matrix $\bH$ to $h(\bx_i)$, where $h(.)$ being any mapping that can well represent the information of a given input $\bx_i$. For example, in linear fitting, $h\left(\bx_i=(\ba_i, b_i)\right)$ can be simply chosen to be $h(\bx_i)=[\ba_i^T \; b_i]$ (i.e., $h$ concatenates $\ba_i$ and $b_i$ to make a row vector). 

Note that since our state involves point sets, one expects our network to be permutation invariant w.r.t. the input $\cS$. In other words, changing positions of the rows in the matrix $\bS_{\cS}$ does not affect the output of our network. Such permutation invariance can be achieved by using the graph CNN architecture, described in the following section. 



\vspace{-2mm}
\subsection{Network (Agent) Design}
\label{sec:agent_design}
\vspace{-1mm}
\begin{figure*}
    \begin{center}
    \includegraphics[width=0.85\linewidth]{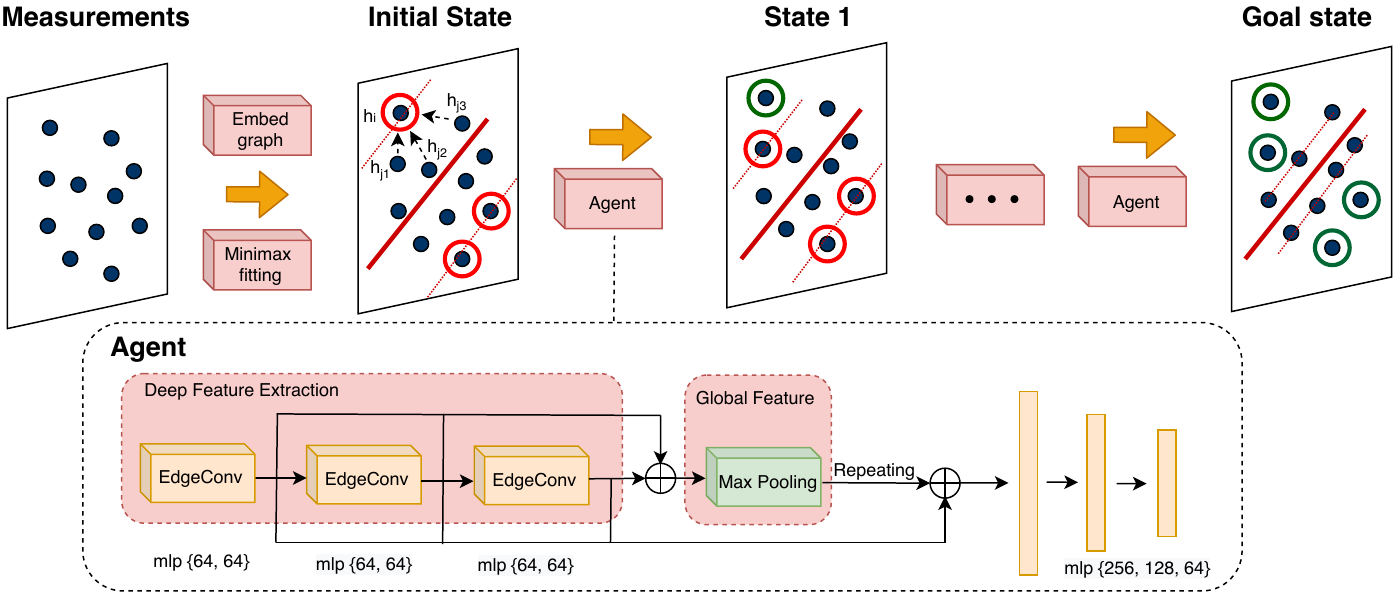}
    \end{center}
       \caption{Illustration of our proposed framework. Top row shows the (unrolled) operations (we use an instance of 2D line fitting as an example). Given a set of measurements, minimax is performed to obtain the initial state. Then, the state is encoded into a graph representation which is fed to the agent to predict the expected Q value (returns) of choosing each point in basis to eliminate. The agent then performs an action, receives a reward and moves to the next state. This process iterates until the agent reaches the goal state. Bottom row depicts the design of our network (agent). }
    \label{fig:mainFramework}
\end{figure*}
Similar to other RL problems, our agent needs to learn to traverse from the initial state to the goal state in such a way that the total rewards is maximized (in the smallest number of steps). This section describes our agent design, which is also illustrated in Figure~\ref{fig:mainFramework}.

The network takes as input a particular state (encoding using the method described in Section~\ref{sec:state_encoding}) and outputs the predicted rewards for the actions associated with the input state.  
We use $\Theta$ to represent the network parameters, we denote by $\hat{Q}(s^{(t)}, a|\Theta)$ the action-value function that returns the optimal reward if the action $a$ is taken given the current state $s^{(t)}$. In other words, the selected action given a current state $s^{(t)}$ is the action that maximizes $\hat{Q}$, i.e., \begin{equation}
    a = \argmax_{a \in \cA(\cB_{\cS})} \hat{Q}(s^{(t)}, a|\Theta),
\end{equation}
where we use $\cA(\cB_{\cS})$ to denote the set of actions associated with the removals of points in the basis set $\cB_{\cS}$ as described in Section~\ref{sec:generating_actions}.

In order to achieve permutation invariance of the input, we design the first stage of our network (shown in the Deep Feature Extraction (DFE) block in Figure~\ref{fig:mainFramework}) to be a series of Edge Convolution (EdgeConv) Layers~\cite{wang2019dynamic}, which was originally inspired from PointNet~\cite{qi2017pointnet}. We employed the EdgeConv layer because, unlike PointNet, it has the ability to capture local geometric structures by taking into account nearest neighbors of each single input (interested readers are referred to~\cite{wang2019dynamic} for more details), hence more information can be extracted to improve the learning capability.
The main role of this DFE block is to capture the relationship between every single input data point and its local geometric structure. Then, the global set feature is obtained from the Global Feature block shown in Figure~\ref{fig:mainFramework}. The repeated concatenation of the global feature with the individual input features is then fed in to a multi-layer perceptron (MLP) to obtain the expected rewards. We then apply a mask to extract only rewards for points in $\cB_{\cS}$.
\vspace{-2mm}
\subsection{Learning algorithm}
\vspace{-1mm}
Based on the components discussed above, this section introduces our general learning algorithm, which is summarized in Algorithm~\ref{alg:MainAlgorithm}. Our learning framework relies on the popular deep Q-learning approach that has been used extensively in several other RL applications.  The training is repeated over multiple episodes. At the start of each episode, a set of measurements $\cX$ containing $N$ data points (with $N$ fixed throughout the training process) is randomly sampled from the training set. Note that the outlier rates in $\cX$ is randomly chosen to be in the range from $1 \%$ to $40 \%$ for each episode.
A minimax fit~\eqref{eq:minimax} is then performed on $\cX$ to obtain the initial state $s_{\cX}$. Starting from $s_{\cX}$, our agent explores the search space by passing through multiple states until reaching the goal state. During the training process, the action taken at each state is sampled based on the popular $\varepsilon$ - greedy policy~\cite{mnih2013playing}. 
After taking an action, the agent receives the reward computed based on~\eqref{eq:reward} and moves to  next state.  The the network parameters are then updated based on the well-known Bellman's equation~\cite{mnih2013playing},
\begin{equation}
    \hat{Q}(s^{(t)}, a|\Theta) = e(s^{t}) + \gamma \max_{a \in \cA(\cB_{\cS})} \hat{Q}(s^{(t+1)}, a|\Theta).
\label{eq:tderror}
\end{equation}
Therefore, the network parameters are updated by minimizing the temporal difference error $\delta$ defined by 
\begin{equation}
     \delta = \rho(e(s^{t}) + \gamma \max_{a \in \cA(\cB_{\cS})} \hat{Q}(s^{(t+1)}, a|\Theta) - \hat{Q}(s^{(t)}, a|\Theta)),
\label{eq:tderror}
\end{equation}
where we choose $\rho$ to be the Huber loss~\cite{huber81}. When the agent reaches the goal state (i.e., $f(\cS) \le \epsilon$), another set of measurements $\cX$ is taken and a new episode is started. As we use the popular PyTorch framework to implement our network, the optimization of~\eqref{eq:tderror} to update the network parameters can be performed by the off-the-shelf gradient-based solvers. More information about the choices of training parameters can be found in the supplementary material. 


\begin{algorithm}[t]
\caption{Main algorithm.}
\begin{algorithmic}[1]
\State Initialize experience relay memory $\cM$
\For {episode e = 1 to L}
\State Take a set of putative measurements $\cX=\{\bx_{i}\}_{i=1}^N$
\State Obtain maximum residual $f(\cS)$ and basis $\cB_{\cS}$ by solving \eqref{eq:minimax}
\State Initialize first state $s^{(t=0)}$
\While{($f(\cS) > \epsilon$)}
    \State
        \begin{equation*}
        \hspace{9mm} a_t= 
        \begin{cases}
            \begin{aligned}
            & \text{random action } a \in \cA(\cB_{\cS}), \;\;  \text{w.p.} \varepsilon  \\
            & \argmax_{a \in \cA(\cB_{\cS})}\hat{Q}(s^{(t)}, a|\Theta), \text{otherwise} \\
            \end{aligned}
        \end{cases}
        \end{equation*}
    \State Get reward $e(s^{t}))$ and move to next state $s^{t+1}$
    \State Add tuple $(s^{t}, a^t, s^{t+1}, e(s^{t}))$ to $\cM$
    \State Sample random batch from $\cM$
    \State Update network parameter $\Theta$
\EndWhile
\EndFor

\end{algorithmic}
\label{alg:MainAlgorithm}
\end{algorithm}

\vspace{-2mm}
\subsection{Local Tree Refinement}
\vspace{-1mm}
Recall from previous sections that the solution returned by our network is a set $\cS^*$ such that $f(\cS^*) \le \epsilon$. Since the proposed algorithm is sub-optimal, the consensus size obtained from $\cS^*$ could be less than the optimal solution $\cI^*$, i.e., $|\cS^*| \le |\cI^*|$,  where $\cI^*$ is the solution of Problem~\eqref{eq:maxcon}. 
To partially overcome this, we propose a simple heuristic in order to gradually improve our obtained solution, which is summarized in Algorithm~\ref{alg:tree_refinement}.
Intuitively, starting from the initial solution $\cS^{t_1 = 0} = \cS^*$ (note that we use $t_1$ to avoid confusion with the state index $t$ used in the previous sections),  we test all points $\bx_j$ in the current outlier set $\hat{\cS}^{t_1} = \cX \setminus \cS^{t_1}$ and add points that lead to consensus size improvement, i.e., $f(\cS^{t_1} \cup \bx_J) \le \epsilon$. This process is repeated until no more points can be added.


\begin{algorithm}[t]
\caption{Local Tree Refinement}
\begin{algorithmic}[1]
\Require Input data $\cX$, initial solution $\cS^{0}$
\State $t \gets 0$, $\texttt{improved} \gets \text{True}$
\While{\texttt{improved} }
    \State $\hat{\cS}^{(t)}\gets \cX \setminus \cS^{(t)}$, $\texttt{improved} \gets \text{False}$
    \For{$\bx_j \in \hat{\cS}^{(t)}$ }
        \If{$f(\cS^{(t)} \cup \bx_j) \le \epsilon$}
        \State $\cS^{(t)} \gets \cS^{(t)} \cup \bx_j$;  $\texttt{improved} \gets \text{True}$. 
        \EndIf
    \EndFor
    \State $t \gets t + 1$.
\EndWhile
\end{algorithmic}
\label{alg:tree_refinement}
\end{algorithm}

\vspace{-1mm}
\section{Experiments}
\vspace{-2mm}
\begin{figure*}
     \centering
     \begin{subfigure}[b]{0.49\textwidth}
         \centering
         \includegraphics[width=1\linewidth]{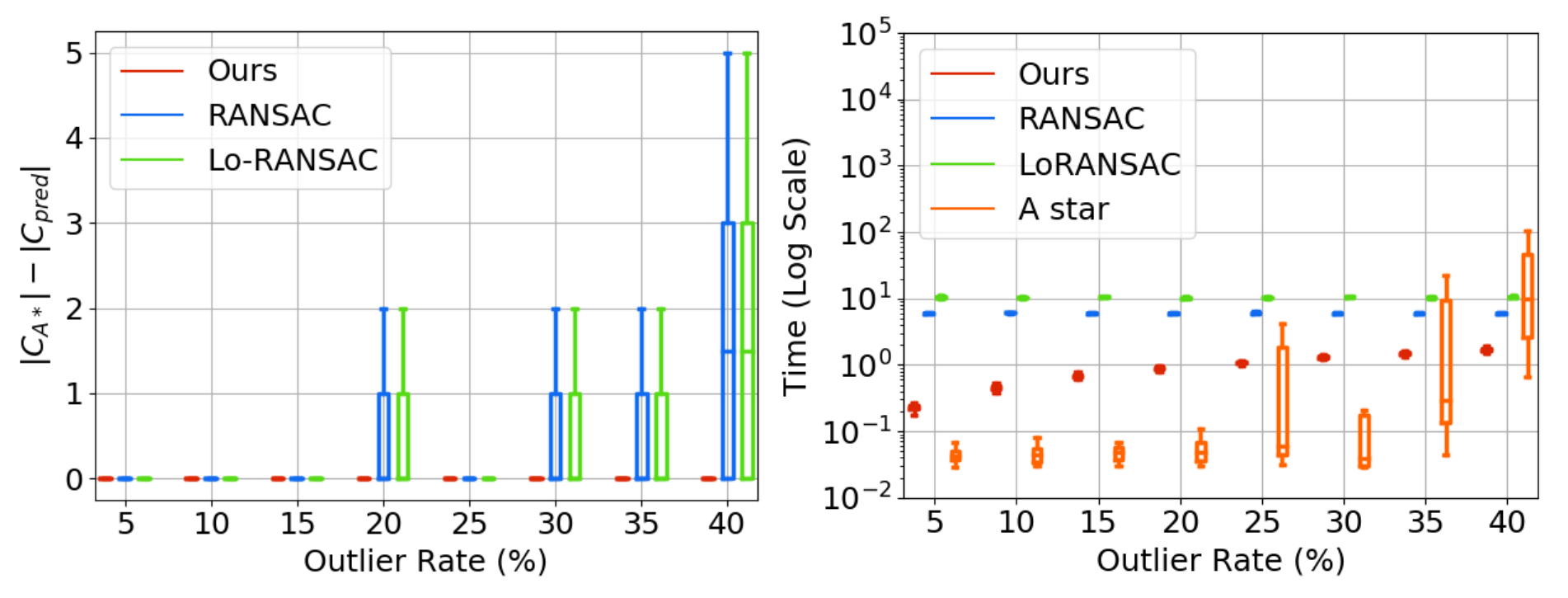}
         \caption{$N = 100$}
         \label{fig:2dresult_100}
     \end{subfigure}
     \hfill
     \begin{subfigure}[b]{0.49\textwidth}
         \centering
         \includegraphics[width=1\linewidth]{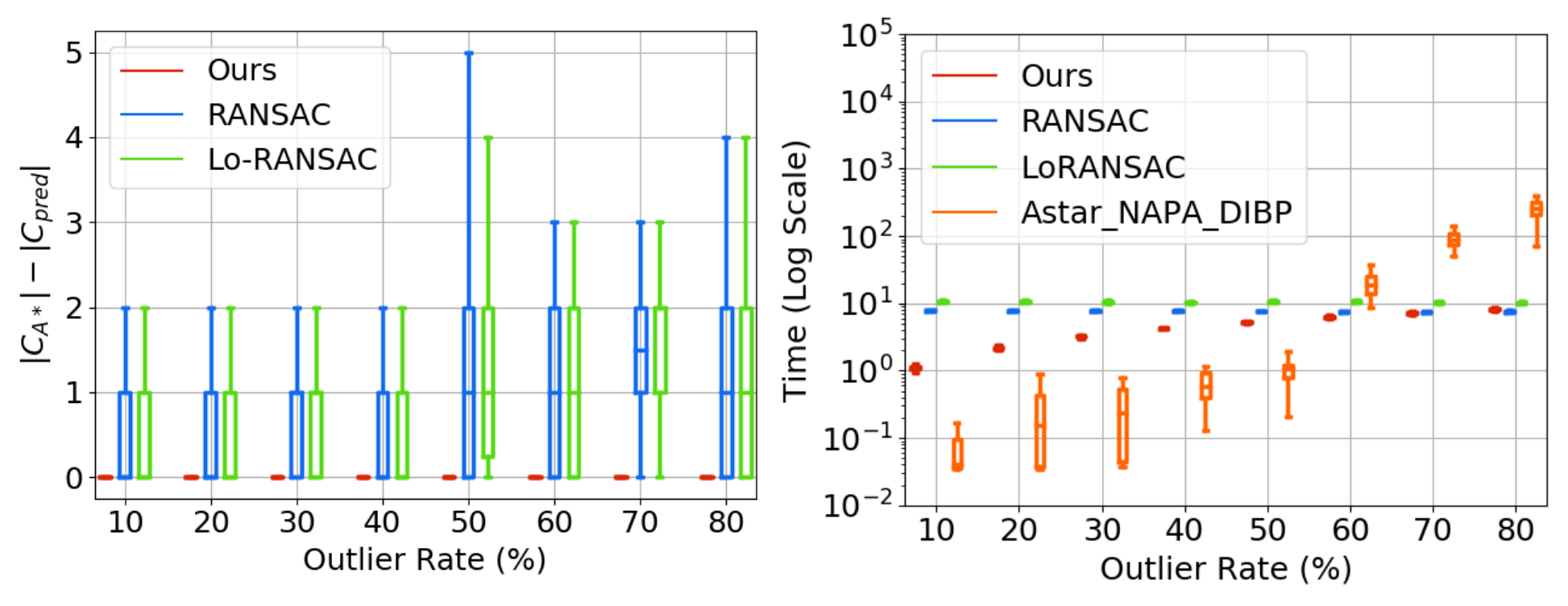}
         \caption{$N = 200$}
         \label{fig:2dresult_200}
     \end{subfigure}
     \vspace{-2mm}
     \caption{2D line fitting on (a) 100 and (b) 200 points, with various outlier rates. Left: Distribution of distance between predicted consensus and global solution (obtained using A*) with baseline models. Right: Run-time(s) of our method compared to optimal methods (A*).}
    \label{fig:2dresult}
    \vspace{-2mm}
\end{figure*}

\begin{figure*}[t]
    \centering
     \begin{subfigure}[b]{0.49\textwidth}
         \centering
         \includegraphics[width=1\linewidth]{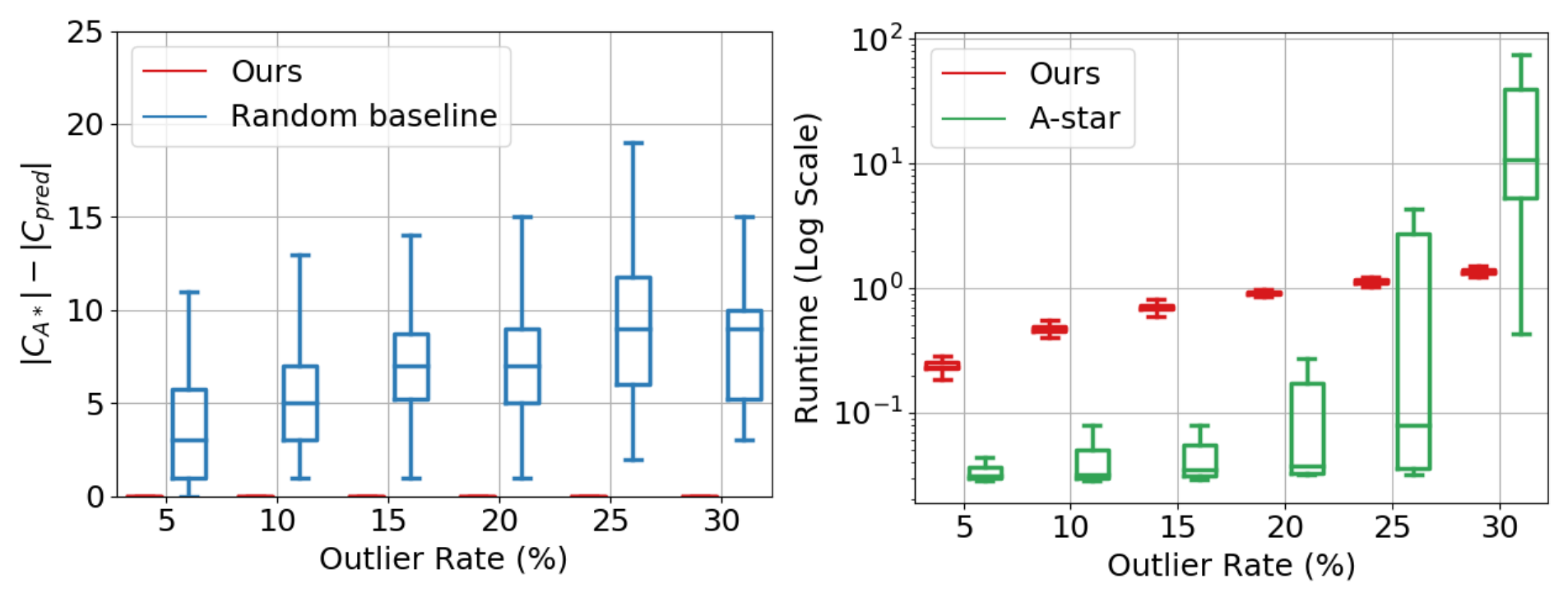}
         \caption{$N = 100$}
         \label{fig:3dresult_100}
     \end{subfigure}
     \hfill
     \begin{subfigure}[b]{0.49\textwidth}
         \centering
         \includegraphics[width=1\linewidth]{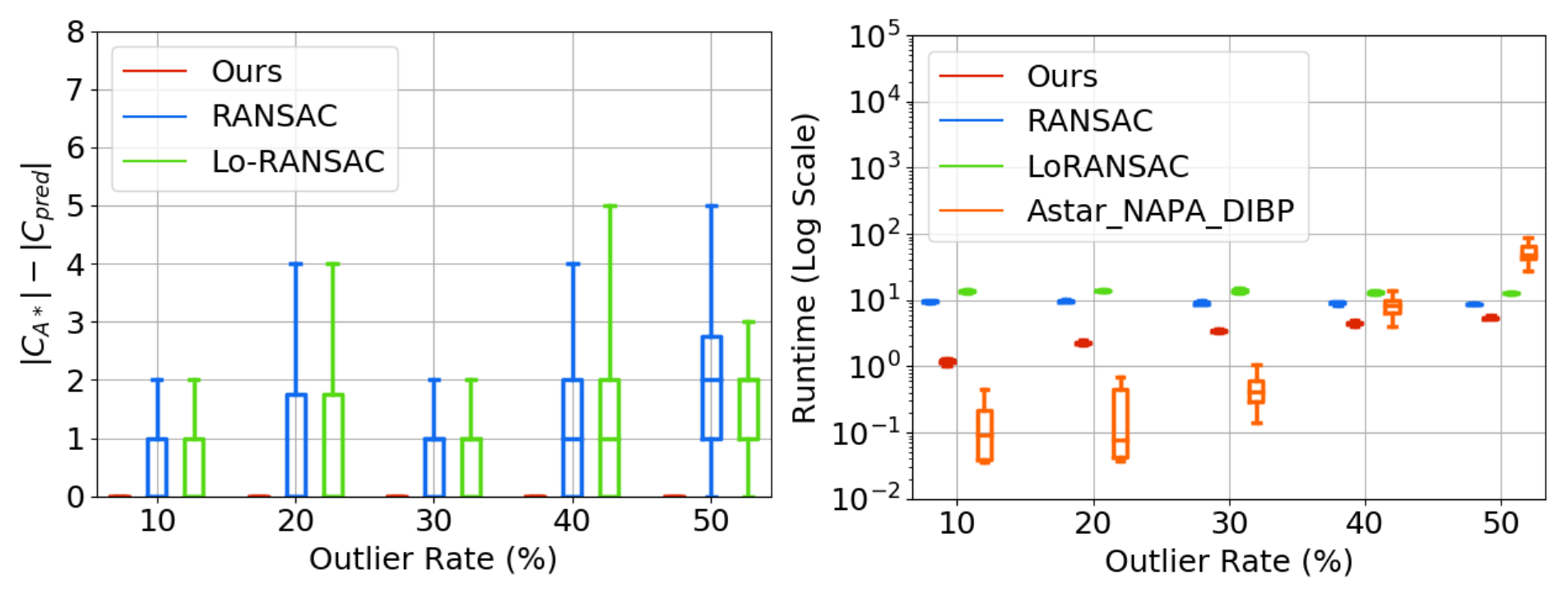}
         \caption{$N = 200$}
         \label{fig:3dresult_200}
     \end{subfigure}
     \vspace{-2mm}
    \caption{3D plane fitting on (a) 100 and (b) 200 points, with various outlier rate. Left: Distribution of distance between predicted consensus and global solution (obtained using A*) with random baseline model. Right: Run-time(s) of our method compared to the optimal method (A*).}
    \label{fig:3dresult}
    \vspace{-3mm}
\end{figure*}

\begin{figure*}[t]
    \centering
    \begin{subfigure}[b]{0.65\textwidth}
         \centering
         \includegraphics[width=1\linewidth]{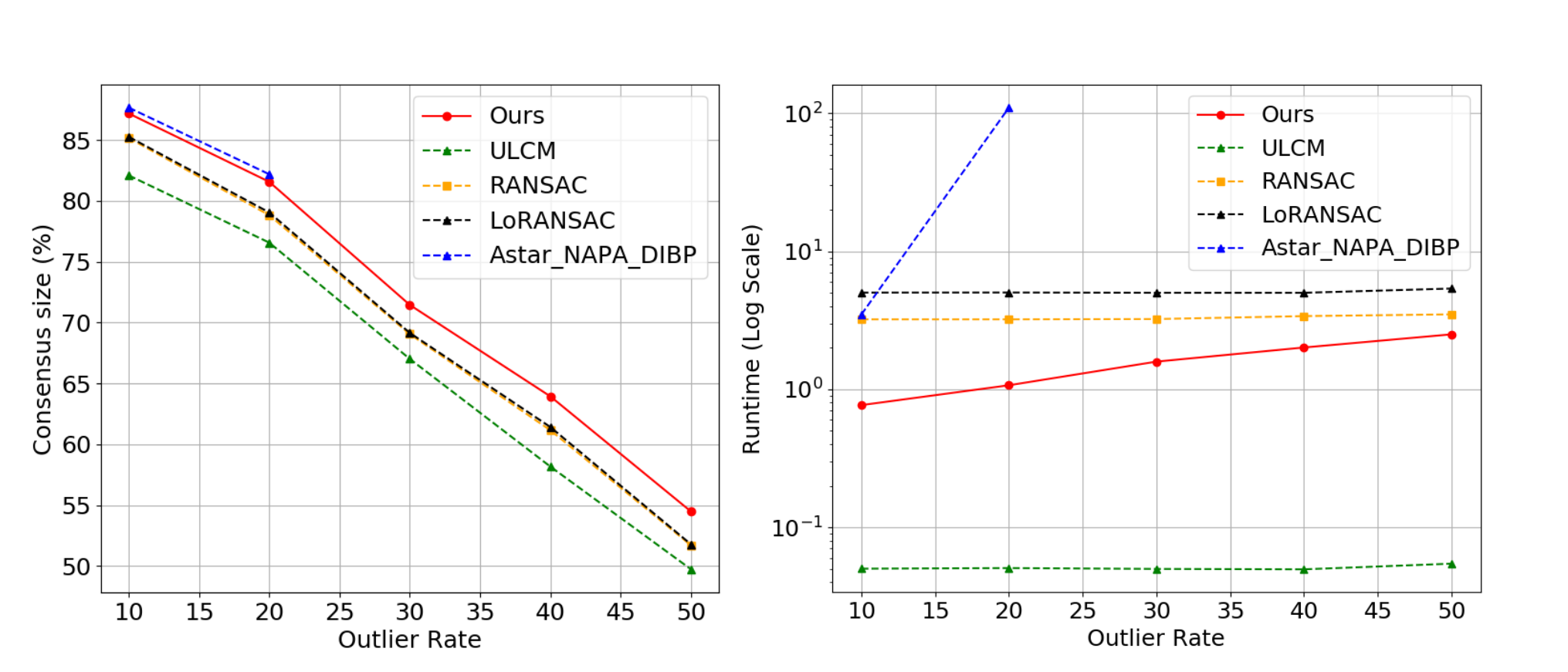}
         \caption{ModelNet40 Dataset. Left: Consensus size comparison. Right: Run time(s) in log scale.}
         \label{fig:fund_synthetic}
    \end{subfigure}
    \hfill
    \begin{subfigure}[b]{0.34\textwidth}
         \centering
         \includegraphics[width=1\linewidth]{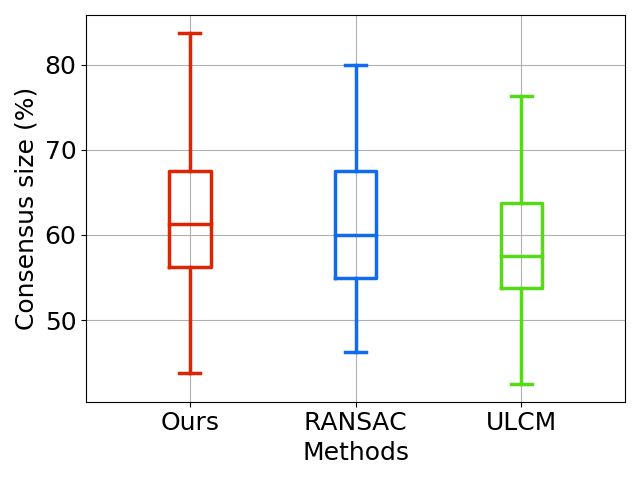}
        \caption{KITTI Dataset.}
         \label{fig:fund_real}
    \end{subfigure}
    \vspace{-3mm}
    \caption{Linearized Fundamental Matrix Estimation with various outlier rates.}
    \label{fig:semisyntheticresult}
    \vspace{-3mm}
\end{figure*}

\begin{figure*}
    \begin{center}
    \includegraphics[width=0.89\linewidth]{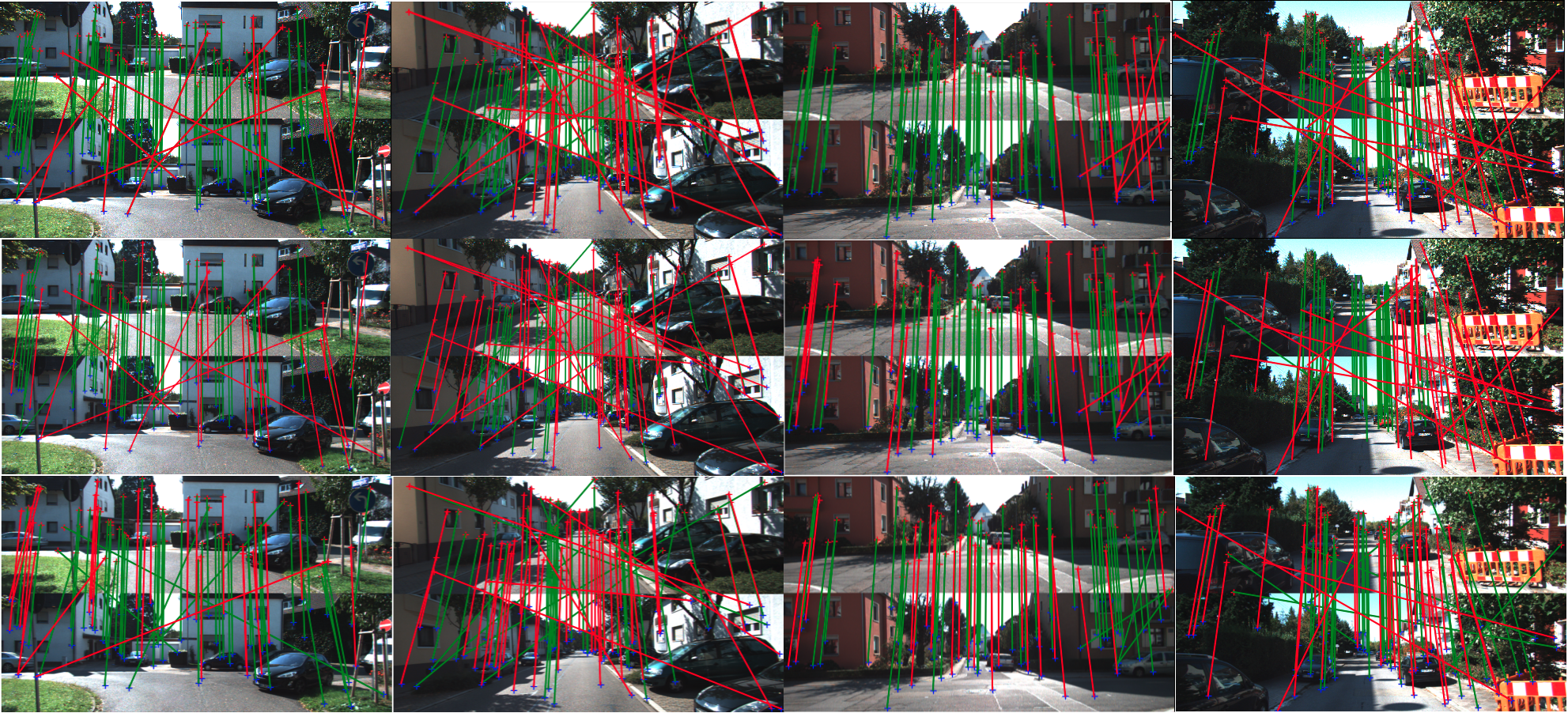}
    \end{center}
    \vspace{-3mm}
       \caption{Qualitative results for robust fundamental matrix. Top: Ours; Middle: RANSAC; Bottom: UCLM}
    \label{fig:realFundamentalQuality}
    \vspace{-3mm}
\end{figure*}

All experiments were executed on an Ubuntu machine with an Intel Core 3.70GHz i7 CPU, 32Gb RAM and a Geforce GTX 1080Ti GPU. Our network is implemented in Python using the popular PyTorch~\cite{paszke2019pytorch} library. We present the main results in this section. A keen reader can refer to the supplementary material for implementation details.

\textbf{Baseline Algorithms.}
We compare our method against the following: Original A* tree search \cite{chin15}, A*- Non-Adjacent Path Avoidance (NAPA) with Dimension Insensitive Branch Pruning (DIBP)~\cite{cai2019consensus}, RANSAC~\cite{RANSAC}, and Local Optimization for RANSAC (LO-RANSAC)~\cite{loransac}. In addition, we also run a random baseline (RB) approach, where at each state, the agent takes an action randomly until reaching a goal state. The objective of comparing with RB is to show that our network traverses the path intelligently as opposed to taking random guesses. 
For fundamental matrix estimation experiments, we also compare our method with the state-of-the-art unsupervised approach for consensus maximization introduced by Probst et al.~\cite{probst2019unsupervised}\footnote{We used the source code provided by the authors with default settings} (since only the source code for fundamental estimation is publicly available, we did not compare against~\cite{probst2019unsupervised} in the line and plane fitting experiments). 
 
In order to have a fair comparison, for each problem instance, we allowed the RANSAC variants to take equal (or longer) run times compared to the run time required by our method.
\vspace{-2mm}
\subsection{Robust Line And Plane Fitting}
\vspace{-1mm}
We first test the algorithms on robust 2D and 3D linear fitting with synthetic data (with various outlier rates) in order to evaluate their performance in a well-controlled setting.
\vspace{-3mm}

\paragraph{Robust 2D Line Fitting.}
To create data (for both training and testing), we randomly generate $N$ (where $N$ is chosen to be $100$ and $200$) points $a_i \in \bbR$, and a line parameter $\hat{\btheta} \in \bbR^2$. From $\{a_i\}$ and $\hat{\btheta}$, we obtain the set $b_i \in \bbR$ by computing $b_i = a_i \theta_1 + \theta_2$. Of the $N$ points in total, $N_o$ are randomly chosen  to be outliers by corrupting their $b_i$ with a uniformly distributed noise between $[-5, -0.1) \cup (0.1, 5]$, while the remaining $N -N_o$ points are perturbed with a uniform noise in the range $[-0.1, 0.1]$. The inlier threshold is set to $\epsilon = 0.1$. When training, we vary the outlier rate from 1 to 40\%. We compare the performance of our network against the random baseline models, and Original A*. Our method and all randomized methods were executed $100$ times. 


We report our results in the form of boxplots to summarize important statistical information, including the median, the variance, and the extreme cases of the data. Since optimal solutions from A* are available in this experiment, we use them as the gold standard to test the quality of the obtained consensus sets. 
The box plot in Fig.~\ref{fig:2dresult_100} (left) depicts the differences in the consensus sizes between A* and other methods for $N=100$ points. 
As shown, our method consistently obtains almost optimal solutions, while the solutions provided by RANSAC and LO-RANSAC are rather unstable over the runs and unsatisfactory for higher outlier rates. 
Fig.~\ref{fig:2dresult_100} (right) shows the run times for all methods. Observe that, compared to A*, our run times are an order-of-magnitude faster. Also note that the run time of A*-NAPA-BIDP~\cite{cai2019consensus} increases exponentially with higher outlier rate. On the other hand, although we allow RANSAC and LO-RANSAC~\cite{loransac} to run longer than ours, we still achieve much better results. We verify the robustness and generalization of our framework by testing it with double the number of points ($N=200$) and higher outlier rates (up to 80\%) without retraining the model. The same conclusions can be drawn from Fig.~\ref{fig:2dresult_200}.
\vspace{-7mm}
\paragraph{Robust 3D Plane Fitting.}

A setting similar to the 2D Fitting experiment (above)  is repeated for 3D plane fitting, which also has many practical applications in computer vision. 
Fig.~\ref{fig:3dresult} shows the performance comparison between our proposed method and a simple random baseline model for $N=100$ and $N=200$. One can see that our proposed method actually learns something, compared to a random method that has ``no knowledge''. We also verify the robustness and generalization of our framework by testing it with double the number of points ($N=200$) and higher outlier rate without retraining the model. 
On average, our proposed method usually achieves optimal solution with a much faster run-time, compared to the A* methods. Moreover, as shown in Fig.~\ref{fig:3dresult}, given a less time budget, our model returns optimal solutions that RANSAC-based approaches never return.
\vspace{-2mm}
\subsection{Robust Fundamental Matrix Estimation}
\vspace{-1mm}
We also test the algorithms on fundamental matrix estimation with linearized residuals, following the settings used in state-of-the-art~\cite{chin15,cai2018deterministic}. For this experiment, we also compare our algorithm with the state-of-the-art unsupervised learning approach ULCM~\cite{probst2019unsupervised}. It is important to highlight the fact that in these experiments it is very difficult to obtain the ground truth as it is impractical to apply the global optimal methods (like A* and its derivatives) since they take took too long to arrive at an optimal solution. Therefore, we use the consensus size as the evaluation metric. 
\vspace{-5mm}
\paragraph{ModelNet40 Dataset}
First, we use data from the ModelNet40~\cite{wu20153d} dataset to train and evaluate our model. Two-view pairs are generated by projecting the point clouds on two image planes having different focal lengths and extrinsic camera parameters. For each image pair, $N=100$ correspondences with outlier rates ranging from $10\%$ to $50\%$ are generated for training and testing. The inlier threshold is chosen as $\epsilon = 0.1$.
The left plot in Fig.~\ref{fig:fund_synthetic} compares the average consensus size on various outlier rate. The run time of A*-NAPA-DIBP~\cite{cai2019consensus} increases exponentially with higher outlier rates. On average, with less time budget, we generally obtain higher consensus sets, compared to RANSAC-based methods. Although the state-of-the-art unsupervised deep learning method, ULCM~\cite{probst2019unsupervised} might be comparatively inexpensive, our method is still able to perform better and achieve higher consensus size. 
\vspace{-5mm}
\paragraph{KITTI Dataset}
We repeat the robust fundamental fitting experiment on one sequence of KITTI dataset \cite{KittiData}. We compute and match SIFT keypoints using the VLFeat toolbox\footnote{http://vlfeat.org}. The inlier threshold is chosen as $\epsilon = 0.01$. Fig.~\ref{fig:fund_real} plots the consensus sizes obtained by our method in comparison with RANSAC and ULCM~\cite{probst2019unsupervised}. 

As seen, our method generally achieves higher consensus sizes compared to ULCM~\cite{probst2019unsupervised}. When compared to RANSAC, although our spread is larger, we have a smaller variance and higher median consensus size. This means that our method only occasionally returns the worst extreme case. We believe that our method can be improved in future work by designing better network architectures to address such outlying scenarios.
We also visualize the solutions qualitatively in Fig.~\ref{fig:realFundamentalQuality} for the three methods. The visualization is consistent with the results shown in Fig.~\ref{fig:fund_real}, where our method achieves inlier sets with higher quality. 



\vspace{-4mm}
\section{Conclusion}
\vspace{-2mm}
We have presented in this work a novel unsupervised reinforcement learning framework for robust estimation. By exploiting the problem structure, we propose an efficient state encoding and a back-bone network that enables our agent to effectively learn to explore the search tree. 
Experimental results show that our method achieves competitive results on several geometric vision problems.
\vspace{-5mm}
\paragraph{Acknowledgments} 
David Suter and Erchuan Zhang acknowledge partial funding for this work under Australian Research Council grant DP200103448.

{\small
\bibliographystyle{ieee_fullname}

}


\end{document}